\documentclass[10pt,twocolumn,letterpaper]{article}

\usepackage[pagenumbers]{cvpr} 

\usepackage[dvipsnames]{xcolor}

\definecolor{cvprblue}{rgb}{0.21,0.49,0.74}
\usepackage[pagebackref,breaklinks,colorlinks,citecolor=cvprblue]{hyperref}

\usepackage{bm}
\usepackage{multirow}
\usepackage{adjustbox}
\usepackage{pifont}
\usepackage{color, colortbl}
\definecolor{Gray}{gray}{0.9}
\usepackage[capitalize]{cleveref}
\crefname{section}{Sect.}{Secs.}
\crefname{table}{Tab.}{Tabs.}

\newcommand{\cmark}{\ding{51}}%
\newcommand{\xmark}{\ding{55}}%
\newcommand{\ieno}{\textit{i}.\textit{e}.}

\def\confName{CVPR}
\def\confYear{2024}

\title{Inter-X: Towards Versatile Human-Human Interaction Analysis}

\newcommand*{\affmark}[1][*]{\textsuperscript{#1}}

\author{
Liang Xu\affmark[1,2] \quad
Xintao Lv\affmark[1] \quad
Yichao Yan\affmark[1*] \quad
Xin Jin\affmark[2*] \quad
Shuwen Wu\affmark[1] \quad
Congsheng Xu\affmark[1] \quad
Yifan Liu\affmark[1] \\
Yizhou Zhou\affmark[3] \quad
Fengyun Rao\affmark[3] \quad
Xingdong Sheng\affmark[4] \quad
Yunhui Liu\affmark[4] \quad
Wenjun Zeng\affmark[2] \quad
Xiaokang Yang\affmark[1]
\vspace{0.7em} \\
\affmark[1]{Shanghai Jiao Tong University} \quad
\affmark[2]{Eastern Institute of Technology, Ningbo} \\
\affmark[3]{WeChat, Tencent Inc.} \quad
\affmark[4]{Lenovo}\\
{\small\url{https://liangxuy.github.io/inter-x/}}
\vspace{-0.5em}
}

\begin{document}

\twocolumn[{
    \renewcommand\twocolumn[1][]{#1}
    \maketitle
    \begin{center}
        \includegraphics[width=1.0\linewidth]{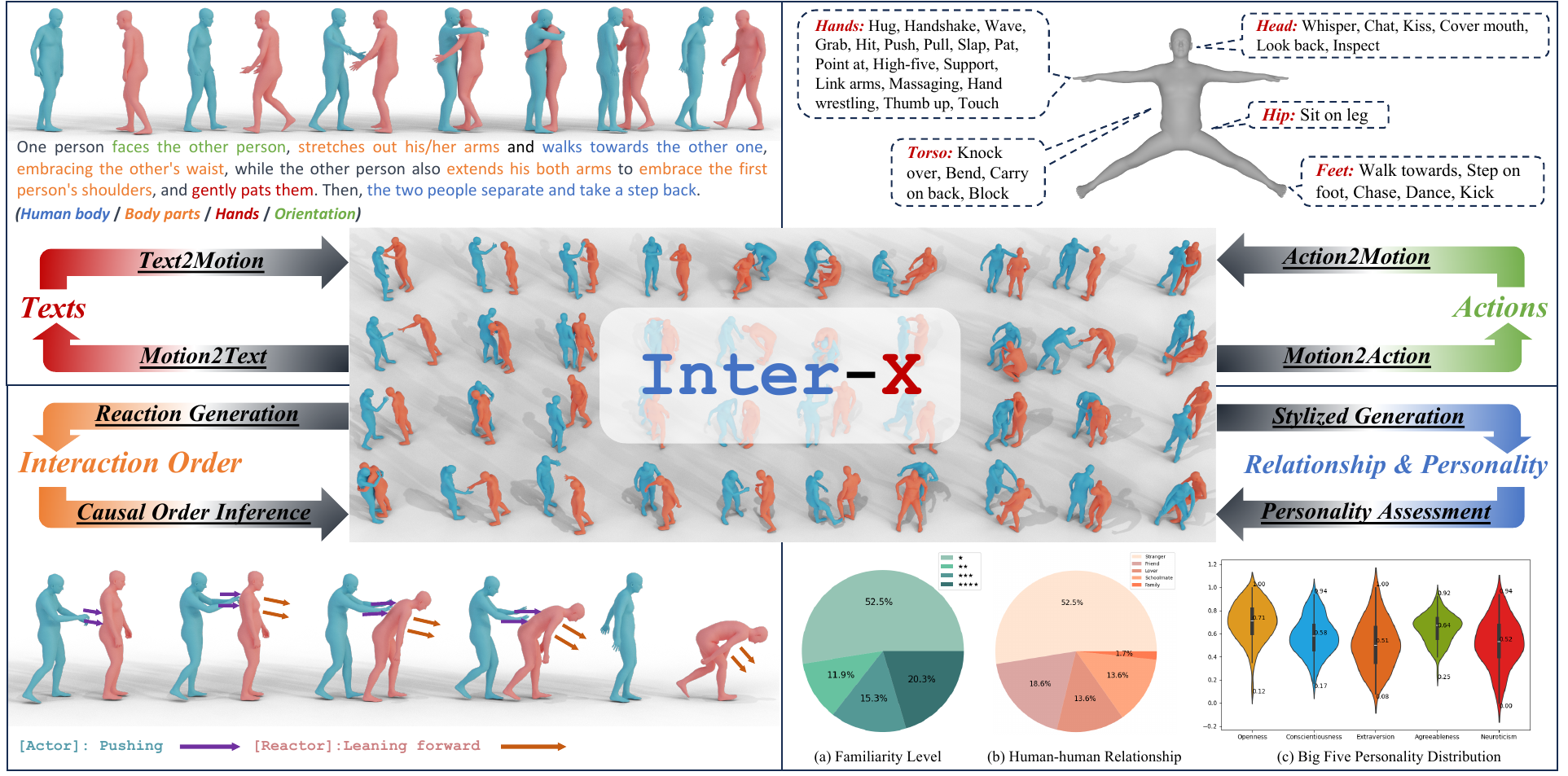}
        \captionof{figure}{An overview of the data and task taxonomy of our proposed Inter-X dataset, which is a large-scale human-human interaction MoCap dataset with $\sim$\textbf{11K} interaction sequences and more than \textbf{8.1M} frames. The fine-grained textual descriptions, semantic action categories, interaction order, and relationship and personality annotations allow for 4 categories of downstream tasks.}
        \label{fig:teaser}
    \end{center}
}]

\let\thefootnote\relax\footnotetext{$^*$Corresponding authors}

\begin{abstract}
\vspace{-5mm}
The analysis of the ubiquitous human-human interactions is pivotal for understanding humans as social beings. 
Existing human-human interaction datasets typically suffer from inaccurate body motions, lack of hand gestures and fine-grained textual descriptions.
To better perceive and generate human-human interactions, we propose Inter-X, a currently largest human-human interaction dataset with accurate body movements and diverse interaction patterns, together with detailed hand gestures. The dataset includes $\sim$\textbf{11K} interaction sequences and more than \textbf{8.1M} frames. We also equip Inter-X with versatile annotations of more than \textbf{34K} fine-grained human part-level textual descriptions, semantic interaction categories, interaction order, and the relationship and personality of the subjects.
Based on the elaborate annotations, we propose a unified benchmark composed of 4 categories of downstream tasks from both the perceptual and generative directions. Extensive experiments and comprehensive analysis show that Inter-X serves as a testbed for promoting the development of versatile human-human interaction analysis.
Our dataset and benchmark will be publicly available for research purposes.
\end{abstract}

\begin{table*}[t]
  \centering
  \begin{tabular}{@{}lccccccccc@{}}
    \toprule
    Dataset & Year & Motions & Frames & Texts & Scheme & Modality & Hands & Asyn. & Rel.\& Pst.\\
    \midrule
    UMPM~\cite{van2011umpm} & 2011 & 36 & 400K & \xmark & MoCap & Skel. & \xmark & \xmark & \xmark\\
    SBU Kinect~\cite{sbu_kinect} & 2012 & 300 & 7.5K & \xmark & RGB+D & Skel. & \xmark & \xmark & \xmark\\
    You2Me~\cite{ng2020you2me} & 2020 & 42 & 77K & \xmark & RGB+D & Skel. & \xmark & \xmark & \xmark\\
    NTU120~\cite{nturgbd120} & 2019 & 8,276 & 462K & \xmark & RGB+D & Skel. & \xmark & \xmark & \xmark\\
    Chi3D~\cite{chi3d} & 2020 & 373 & 63K & \xmark & MoCap & SMPL-X & \cmark & \xmark & \xmark\\
    ExPI~\cite{expi} & 2022 & 115 & 30K & \xmark & mRGB & Skel. & \xmark & \xmark & \xmark \\
    Hi4D~\cite{hi4d} & 2023 & 100 & 11K & \xmark & mRGB & SMPL & \xmark & \xmark & \xmark \\
    InterHuman~\cite{interhuman} & 2023 & 6,022 & 1.7M & 16,756 & mRGB & SMPL & \xmark & \xmark & \xmark \\
    \midrule
    Inter-X & 2023 & \textbf{11,388} & \textbf{8.1M} & \textbf{34,164} & MoCap & SMPL-X & \cmark & \cmark & \cmark\\
    \bottomrule
  \end{tabular}
  \caption{\textbf{Dataset comparisons.} We compare our Inter-X dataset with the existing human-human interaction datasets. \textbf{Motions}: The number of the motion clips; \textbf{Frames}: The frame number of the 3D human motions; \textbf{Texts}: The number of the textual descriptions; \textbf{Scheme}: The strategy to obtain the motion data; \textbf{Modality}: The representation of the motion data and ``Skel.'' denotes skeleton; \textbf{Hands}, \textbf{Asyn.} and \textbf{Rel.\&Pst.} refer to the components of hand gestures, asymmetry annotations, human-human relationships and personalities.}
  \label{tab:dataset}
\end{table*}

\vspace{-7mm}
\section{Introduction}

The ability to perceive and generate human-human interactions is fundamental in constructing intelligent digital human systems, which have numerous applications in surveillance, AR/VR, games, and robotics.
However, this task is challenging due to the complex and diverse interaction patterns, as well as self-occlusions.
Although impressive progress has been made in the perception tasks, \ieno, skeleton-based interaction recognition~\cite{vahdat2011discriminative,ji2014interactive,perez2021interaction,pang2022igformer,duan2023skeletr}, and the generation tasks, \ieno, action/text-conditioned interaction generation~\cite{humanml3d,xu2023actformer,actor,mdm,interhuman}, they remain sub-optimal due to the lack of a comprehensive dataset to cover all the aspects of this task.

The advancement of human-human interaction analysis is accompanied by the construction of human-human interaction datasets~\cite{van2011umpm,sbu_kinect,ng2020you2me,chi3d,nturgbd120,expi,hi4d,interhuman}, as listed in~\cref{tab:dataset}. However, we believe that all the previous datasets remain unsatisfactory on the following aspects: 
1) \textbf{Expressive ability,} \ieno, the dexterous hand gestures play important roles for human-human interactions, like ``shaking hands'', ``grabbing'', ``waving'', \etc. However, to the best of our knowledge, there is no large-scale dataset providing high-fidelity finger movements for human-human interactions.
2) \textbf{Fine-grained text descriptions,} \ieno, text-driven generative tasks are promising for practical applications and have attracted much attention. Unlike coarse text annotations like ``one person approaches the other and embraces her/him'', fine-grained descriptions with human part-level semantics enable controllable interaction generation and better alignment~\cite{action-gpt} between motion and text modalities, spatiotemporally.
3) \textbf{Interaction order,} \ieno, during a causal human-human interaction period such as ``kicking'', the actor and reactor are asymmetric. However, the asymmetry property for human-human interactions is not considered in previous datasets.
4) \textbf{Relationship and personality,} \ieno, the intimacy level and social relationships between individuals together with their personalities intuitively affect the interaction patterns, which should be considered.

To address the aforementioned limitations of existing datasets, we thus build a large-scale human-human interaction dataset, called Inter-X, as depicted in~\cref{fig:teaser}, with precise, diverse human-human interaction sequences, and detailed hand gestures. 
To capture Inter-X, we first build a MoCap system with the combination of the optical scheme to capture accurate body movement and the inertial solution to record hand gestures against occlusion.
Inter-X covers 40 daily interaction categories, $\sim$\textbf{11K} motion sequences with more than \textbf{8.1M} frames. We recruited 89 distinct subjects with different social relationships, \ieno, strangers, friends, lovers, schoolmates, and family members. We also collect their familiarity levels and their individual Big Five personalities~\cite{wiggins1996five,vinciarelli2014survey,delgado2022automatic}.

With our proposed high-precision human-human interaction dataset and the versatile annotations, as illustrated in~\cref{fig:teaser}, we empower 4 categories of downstream tasks with half of them as generative tasks and the remaining as perceptive tasks. 
1) \textbf{Texts} enable not only controllable human interaction generation from natural languages~\cite{interhuman} but also the human interaction captioning tasks~\cite{guo2022tm2t,jiang2023motiongpt};
2) \textbf{Action categories} facilitate action-conditioned human interaction generation~\cite{xu2023actformer} together with the human interaction recognition tasks~\cite{pang2022igformer,duan2023skeletr};
3) \textbf{Interaction order} enables the causal human reaction generation~\cite{chopin2023interaction,role_aware,ghosh2023remos,liu2023interactive} and the causal order inference tasks, \ieno, detecting the perpetrator in surveillance scenarios;
4) \textbf{Relationship and personality} make the stylized interaction generation~\cite{aberman2020unpaired,jang2022motion} and the personality assessment possible.
We formulate our Inter-X dataset as a unified testing ground for all the downstream tasks.
For the existing tasks, we extensively evaluate the state-of-the-art methods on the Inter-X's test set with extensive discussions.
We also build up the baseline methods and evaluation metrics for the remaining tasks.

In summary, our contributions can be summarized as follows:
1) We collect the currently largest human-human interaction dataset with accurate human body movements, diverse interaction patterns, and expressive hand gestures;
2) We complement Inter-X with fine-grained human part-level textual descriptions, semantic action categories, causal interaction order annotations, relationship and personality information.
3) We propose a unified human-human interaction benchmark with 4 categories of downstream tasks to enable extensive research directions.

\section{Related work}

\subsection{Human motion datasets}
Compared to RGB videos, human motion representation is high-level, efficient, privacy-friendly and robust to illumination~\cite{nturgbd120,xu2022skeleton}. Human motion datasets with action labels~\cite{nturgbd120,humanact12,uestc,punnakkal2021babel} and text descriptions~\cite{plappert2016kit,humanml3d,motion-x} facilitate the development for understanding human motions. Datasets accompanied with audio signals~\cite{li2021learn,Valle-Perez2021Transflower} and scene/object conditions~\cite{taheri2020grab,hassan2019resolving,zhang2022couch,hassan2021stochastic,araujo2023circle,wang2022humanise} are also produced for real-world human-centric tasks.

\subsection{Human-human interaction datasets}
Besides the single-human motion datasets, many human-human interaction datasets have been proposed~\cite{van2011umpm,sbu_kinect,ng2020you2me,chi3d,nturgbd120,expi,hi4d,interhuman} as listed in~\cref{tab:dataset} with various sizes, modalities and functionalities. Especially, InterHuman~\cite{interhuman} was recently built as a large-scale human-human interaction dataset with textual annotations.
However, as aforementioned, our Inter-X dataset still maintains advantages with respect to motion quality, fine-grained textual annotation, detailed hand gestures, and comprehensive annotation modalities.

\subsection{Perceptive tasks for human motion}
Skeleton-based human action recognition has been a long-standing problem for years~\cite{stgcn,2s-agcn,ms-g3d,semantics2019,chengdecoupling,zhang2020context,ctrgcn,zhou2023learning,foo2023unified,lee2023hierarchically,hdgcn}. Compared to it, human interaction recognition~\cite{vahdat2011discriminative,ji2014interactive,perez2021interaction,pang2022igformer,duan2023skeletr} is a sub-field of it, relying on modeling the semantic correlations between humans.
Besides human action recognition, human motions contain biometric cues about human subjects~\cite{deligianni2019emotions,vinciarelli2014survey}. Gait recognition~\cite{wan2018survey,sepas2022deep} aims to identify the individuals from human motions. Other works like\cite{durupinar2016perform,delgado2022automatic} regard the human movements as personality predictors.
Our Inter-X dataset with large-scale action-motion and text-motion pairs will promote the development of human action recognition. We also take a significant step forward in assessing the human-human relationships and personalities from human motions.

\subsection{Generative tasks for human motion}
The goal of human motion generation is to generate plausible and diverse motion data based on different guidances. Human motion generation from action labels~\cite{action2motion,actor,mdm,xu2023actformer,cervantes2022implicit,chen2022executing}, textual descriptions~\cite{language2pose,motiondiffuse,petrovich22temos,guo2022generating,flame,mofusion,motion-x,lu2023humantomato} and audios~\cite{dance2music,aristidou2021rhythm,li2022danceformer,habibie2022motion,ao2022rhythmic,Ao2023GestureDiffuCLIP} have emerged in recent years. Besides single-person human motion generation,~\cite{xu2023actformer,commdm,interhuman} attempt to generate multi-person interactions. Besides, a few works~\cite{chopin2023interaction,role_aware} tackle the problem of generating the reaction between two interactions. 
To enhance the expressibility of the generated motions,~\cite{aberman2020unpaired,jang2022motion} manage to solve motion style transfer and stylized motion generation tasks.
Our Inter-X dataset can be utilized for action or text-conditioned human interaction generation tasks. The explicit interaction order annotations greatly facilitate the reaction generation task. At the same time, personalities and relationships can serve as factors for stylized human interaction generation.

\subsection{Multimodality in vision}
The world surrounding us involves multiple modalities~\cite{baltruvsaitis2018multimodal,guo2019deep,xu2023multimodal}, so are the ubiquitous human-human interactions. Many multimodal datasets~\cite{tang2023flag3d,hi4d,interhuman,motion-x} related to human motions emerged in recent years. Based on our multimodal Inter-X dataset, we unify several categories of downstream tasks towards a deeper understanding of human-human interactions.

\section{The Inter-X Dataset}

We present the large-scale Inter-X dataset towards versatile human-human interaction analysis, which consists of 11,388 interaction sequences and more than 8.1M frames, covering 40 daily interaction categories and 89 subjects.

\subsection{Data Capturing System}

\begin{figure}[t]
  \centering
   \includegraphics[width=1.0\linewidth]{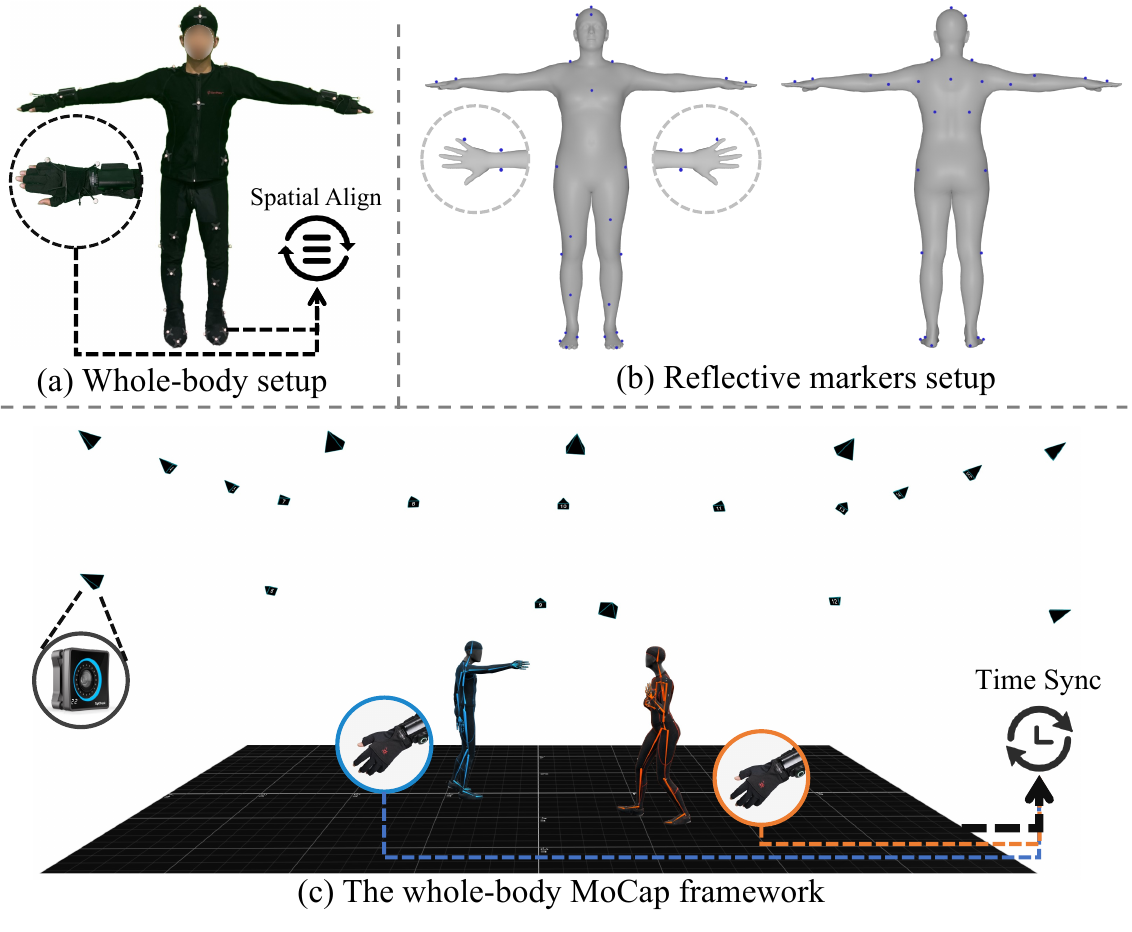}
   \caption{\textbf{An overview of the Inter-X capture system.} (a). The optical MoCap clothing together with the inertial gloves are spatially integrated via a triangular bracket of reflective markers; (b). The details of the markers setup; (c). The body and hands are temporally synchronized in the whole-body MoCap framework.}
   \label{fig:setup}
\end{figure}

Most of the previous datasets take the multi-view RGB-based technologies~\cite{nturgbd120,interhuman}, \ieno, extracting the human motion from RGB videos. Though the natural RGB images are captured, these datasets suffer from severe occlusions and penetrations, and the subtle finger movements are hard to obtain precisely.
For the trade-off between accuracy and natural RGB images~\cite{taheri2020grab}, we prioritize accuracy and thus choose the optical MoCap system for body movements. Additionally, we adopt inertial gloves to capture the finger gestures, which are robust to occlusions. The overview of our capturing system is illustrated in~\cref{fig:setup}.

The length, width, and height of our MoCap venue are 8.5 meters, 5.4 meters, and 3.3 meters, which is capable of covering most daily human-human interactions. We deploy the OptiTrack MoCap system~\cite{optitrack} with 20 PrimeX 22 infrared cameras. For each camera, we capture the resolution of 2048$\times$1088 at 120 fps. 
The optical motion capture scheme ensures a $\pm0.15$mm error, much lower than the RGB camera scheme.

To capture the dexterous hand gestures without occlusion, we adopt the inertial solution of the commercial Noitom Perception Neuron Studio (PNS) gloves~\cite{noitom}. The subtle finger movements can be captured in real-time, disregarding the self-occlusion and occlusion with the other person during the interactions. We also re-calibrate the PNS gloves frequently to mitigate the error accumulation.

For each group of two volunteers, they wear the MoCap suits with 41 reflective markers and the inertial gloves as depicted in~\cref{fig:setup}(a),(b). Both of them are carefully calibrated before they perform the interactions. We provide timecodes for the OptiTrack MoCap system and the PNS gloves so that the body and hands can be temporally synchronized. For each batch of the shoot, we arrange five action categories with five repetitions for variability, which improves efficiency and also ensures the continuity of the volunteers' actions. The volunteers pause for several seconds between two interaction snippets to ease the subsequent segmentation.
More details of the data capturing processing can be found in the supplementary materials.

\subsection{Data Postprocessing}
The crux of the postprocessing is the alignment between the body poses from the OptiTrack MoCap system and the finger gestures from the inertial gloves. Temporally, we retrieve the intersection of the body pose and hand pose sequences. Spatially, they are naturally integrated through the shared wrist rotation from the triangular locating bracket. Given the spatiotemporally aligned motion sequences, the annotators should segment the start and end frames for each atomic interaction snippet. We collect, check the temporal segmentation results, and then trim the long recorded motion sequences into atomic segments.

\section{Dataset Taxonomy}

\begin{figure}[t]
  \centering
   \includegraphics[width=1.0\linewidth]{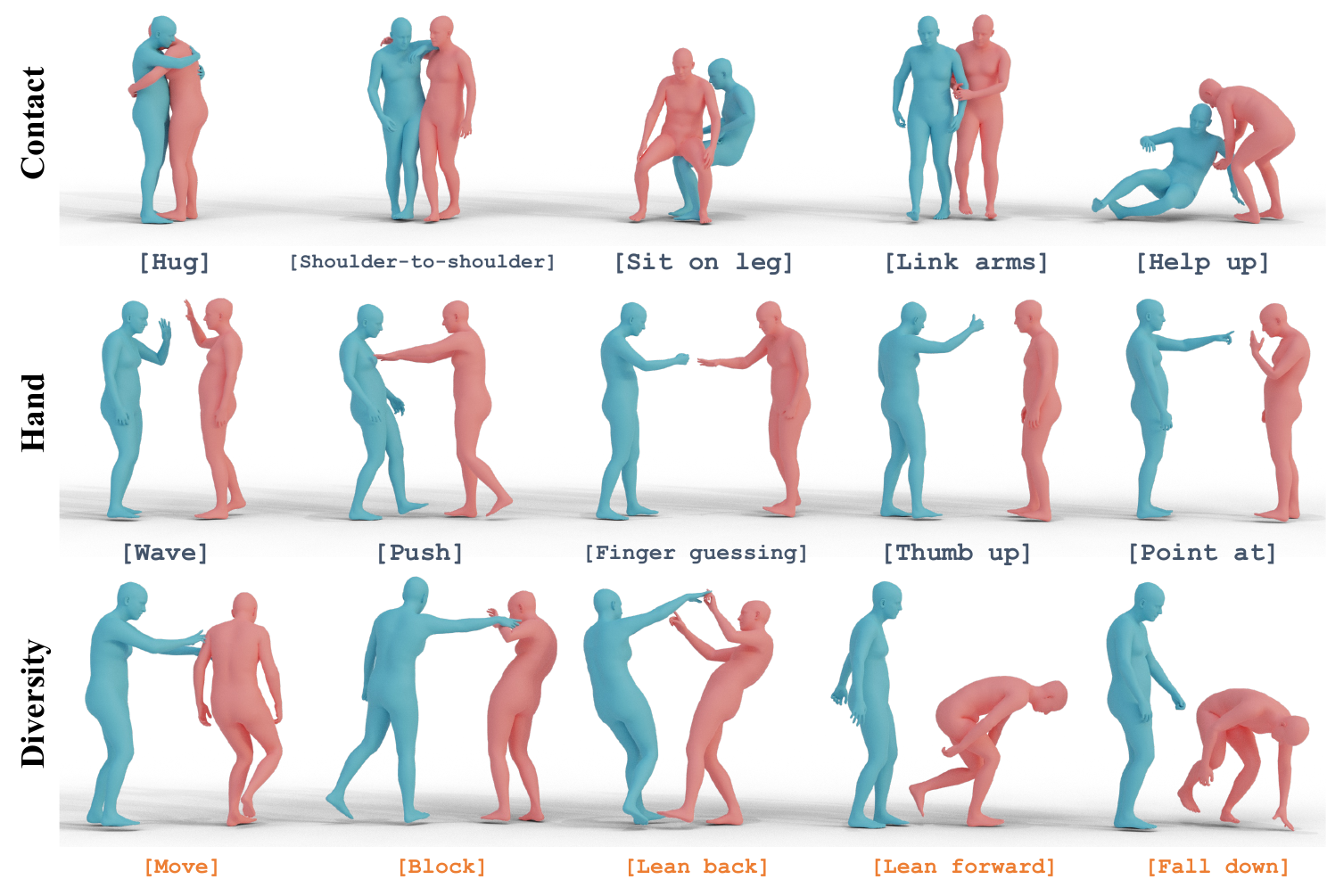}
   \caption{\textbf{More examples of the Inter-X dataset.} Our proposed Inter-X dataset for human-human interaction analysis is highly accurate, hand gestures incorporated, with diverse actions and reactions. Please zoom in for the details.}
   \label{fig:dataset}
\end{figure}

We enrich the high-precision human-human interaction sequences with multifaceted modalities, resulting in 13,888 pairs of SMPL-X~\cite{smplx} motion sequences, 273,312 synthetic multi-view RGB videos, 34,164 detailed text descriptions, 40 semantic action categories with diverse action/reaction patterns, interaction order labels, and the relationship for 59 groups and personality for 89 volunteers. \cref{fig:dataset} shows some characteristics of the Inter-X dataset.

\subsection{Interaction data}
\noindent{\textbf{MoCap Data.}}
We adopt the SMPL-X parametric model for its expressivity for human body poses and articulated hand poses, and the generality for various downstream tasks. Formally, the SMPL-X parameter is composed of the body pose parameters $\theta\in\mathbb{R}^{N\times55\times3}$, shape parameters $\beta\in\mathbb{R}^{N\times10}$ and the translation parameters $t\in\mathbb{R}^{N\times3}$, where $N$ is the number of the frames. We initialize the shape parameters $\beta$ based on the height and the weight of the volunteer as~\cite{virtual_caliper}. Then an optimization algorithm is well-tuned to fit the SMPL-X parameters based on the captured key points:
\begin{equation}
  E(\theta,t)=\lambda_1\frac{1}{N}\sum_{j \in \mathcal{J}}\lambda_{p}||\bm{J}_j(\mathbb{M}(\theta,t))-\bm{g}_j||_2^2+\lambda_2||\theta||_2^2,
  \nonumber
\end{equation}
where $\mathcal{J}$ denotes the joints set, $\mathbb{M}$ is the SMPL-X parametric model, $\bm{J}_j$ is the joint regressor function for joint $j$, $\bm{g}$ is the skeleton captured from the MoCap system. $\lambda_1$, $\lambda_2$ and $\lambda_p$ are different weights and we apply different weights for different body parts. Please refer to the supplementary materials for more details.

\noindent{\textbf{Rendered RGB.}}
The synthetic data has broad applications for human motions~\cite{fabbri2018learning,gta-human,xu2023actformer,black2023bedlam}. To enrich our Inter-X dataset with RGB modality, we utilize the Unreal Engine to render multi-view 2D videos similar to~\cite{tang2023flag3d}. We download the free character models from Renderpeople~\cite{renderpeople}, and then retarget our full-body interaction data to the rigged characters. 
We select the realistic scene models from the Unreal Engine Store and then place the Renderpeople models into them. We capture multi-view videos with 6 rounded cameras, with a resolution of 1920$\times$1080 and a frame rate of 30 fps. Ultimately, 273,312 synthesized RGB videos with 11,388 interaction sequences, 4 different scenes and 6 viewpoints are generated.

\subsection{Action categories}
We choose the action categories referring to the existing human-human interaction datasets~\cite{chi3d,nturgbd120,interhuman} and large language models~\cite{gpt3}. Finally, we figure out 40 daily human-human interaction categories, which cover the most interaction categories to the best of our knowledge.
We instruct each volunteer to perform \textit{naturally} and \textit{diversely}. For diversity, the volunteers can perform 1) Diverse actions, \ieno, raising left hand, right hand, or both hands when ``raising hands''; 2) Diverse reactions, \ieno, rebelling, taking a few steps back or falling down when being ``pushed''; 3) Diverse human boy states, \ieno, standing, sitting, crouching or even lying on the ground.
Each interaction is repeated five times for variability.

\subsection{Text descriptions}

Textual descriptions, especially fine-grained ones, empower various practical applications for better perception and generation. We implement an annotation tool based on~\cite{ait-viewer}, so that the annotators can scale and rotate the view for 360 degrees to observe the details of the interactions. For each interaction sequence, we ask 3 distinct annotators to describe it from human part levels with 1) the coarse body movements, 2) the finger movements, and 3) the relative orientations.
We correct the typos of the collected textual descriptions with GPT-3.5~\cite{gpt3} and then spot-check the results. Upon analysis, the average length of our textual descriptions is $\sim$35, which significantly surpasses existing action datasets, reflecting the fine-grained nature of our texts.

\subsection{Interaction Order}
The study of causal relationships, where one person acts and the other one reacts, could help extend the understanding of human-human interactions~\cite{sbu_kinect}. We ask the volunteers to explicitly annotate the order of the actors and reactors for each atomic interaction sequence.

\subsection{Relationship \& Personality}

Exploring the correspondence between human motion and personality is a niche~\cite{durupinar2016perform,delgado2022automatic}, and the essence lies in the disentanglement of the personality factors from motions. We adopt the dominant paradigm of the Big-Five Personality Model~\cite{wiggins1996five,vinciarelli2014survey,delgado2022automatic}. The participants are asked to fill out the NEO Five-Factor Inventory~\cite{mccrae2004contemplated} to measure their personalities from openness, conscientiousness, extraversion, agreeableness and neuroticism perspectives. Besides, the volunteers fill out the questionnaire to rank their familiarity level from levels 1 to 4, and declare their social relationships of 5 categories, \ieno, strangers, friends, lovers, schoolmates, and family.

\section{Task Taxonomy}

Our high-precision human-human interaction MoCap data with dexterous hand details bring vitality and challenge to existing tasks. Moreover, we also propose different downstream tasks with practical applications tailored to the versatile annotations. Formally, we denote each human-human interaction sequence as $\bm{m}$$=$$<$$\bm{x}, \bm{y}$$>$, and the annotations as action category $\bm{l}_a$, text description $\bm{l}_t$, causal interaction order $\bm{l}_c$, relationship $\bm{l}_r$ and personalities $\bm{l}_p$$=$$<$$\bm{l}_{p_x}, \bm{l}_{p_y}$$>$.

\subsection{Texts related Tasks}

\noindent{\textbf{Text-conditioned human interaction generation.}} Text-conditioned single-person human motion generation has been widely explored with various datasets~\cite{plappert2016kit,humanml3d,motion-x} and models. We pose opportunities for controllable human-human interaction generation~\cite{motion-x,action-gpt} with fine-grained textual annotations and challenges to synthesize the subtle hand gestures and the alignment between human part-level textual descriptions and interactions. The task can be represented as learning a function $F_{t2m}$:
\begin{equation}
  F_{t2m}(\bm{l}_t) \mapsto \bm{m}.
  \label{eq:t2m}
\end{equation}

\noindent{\textbf{Human interaction captioning.}} Human interaction captioning is a newly proposed task~\cite{guo2022tm2t,jiang2023motiongpt}, to generate corresponding textual descriptions rather than recognizing the action category given a human-human interaction sequence, which can boost the alignment between texts and motion data and automatically generate diverse and reasonable textual descriptions. This task can be formulated as:
\begin{equation}
  F_{m2t}(\bm{m}) \mapsto \bm{l}_t.
  \label{eq:m2t}
\end{equation}

\subsection{Actions related Tasks}

\noindent{\textbf{Action-conditioned human interaction generation}.} Given an action label, $F_{a2m}(\cdot)$ aims to generate diverse and plausible human-human interaction sequences~\cite{xu2023actformer,actor,mdm}. With our proposed Inter-X, we can generate more realistic and detailed interactions with fingers:
\begin{equation}
  F_{a2m}(\bm{l}_a) \mapsto \bm{m}.
  \label{eq:a2m}
\end{equation}

\noindent{\textbf{Human interaction recognition}.} Human interaction recognition has practical applications for visual surveillance~\cite{pang2022igformer,duan2023skeletr}. We believe that integrating the fine hand movements will enhance the recognition ability of current models. We formulate this task as:
\begin{equation}
  F_{m2a}(\bm{m}) \mapsto \bm{l}_a.
  \label{eq:m2a}
\end{equation}

\subsection{Interaction-order related Tasks}

\noindent{\textbf{Human reaction generation}.} Human reaction generation~\cite{chopin2023interaction,role_aware,ghosh2023remos,liu2023interactive} is a less explored problem yet with broad applications in AR/VR and gaming. Explicit annotations of the actor-reactor order will advance the research on the asymmetry of different roles with human-human interactions:
\begin{equation}
  F_{c2m}(\bm{l}_c, \bm{x}) \mapsto \bm{y}.
  \label{eq:c2m}
\end{equation}

\noindent{\textbf{Causal order inference}.} $F_{m2c}(\cdot)$ aims to differentiate the actor and reactor given a human interaction sequence, which will benefit intelligent surveillance and sports:
\begin{equation}
  F_{m2c}(\bm{m}) \mapsto \bm{l}_c.
  \label{eq:m2c}
\end{equation}

\subsection{Relationship \& Personality related Tasks}

\noindent{\textbf{Stylized human interaction generation}.} The relationship between two participants and their personalities can serve as stylization factors for customized human interaction generation. The large number of participants with each having a long sequence of motion data enable us to accomplish this task. We formulate this task as:
\begin{equation}
  F_{s2m}(\bm{l}_a, \bm{l}_r, \bm{l}_p) \mapsto \bm{m}.
  \label{eq:s2m}
\end{equation}

\noindent{\textbf{Personality assessment}.} Previous works~\cite{durupinar2016perform,delgado2022automatic} regard the body movements of participants as personality predictors. Leveraging our Inter-X dataset, we propose a new task of personality and relationship assessment, which is vital for education, medicine, sports, \etc. Specifically,
\begin{equation}
  F_{m2s}(\bm{m}) \mapsto \{\bm{l}_r, \bm{l}_p\}.
  \label{eq:m2s}
\end{equation}

\begin{table*}[t]
  \centering
  \resizebox{1\textwidth}{!}{
  \begin{tabular}{@{}lccccccc@{}}
    \toprule
    \multirow{2}{*}{Methods}  & \multicolumn{3}{c}{R Precision$\uparrow$} & \multirow{2}{*}{FID $\downarrow$} & \multirow{2}{*}{MM Dist$\downarrow$}  & \multirow{2}{*}{Diversity$\rightarrow $} & \multirow{2}{*}{MModality $\uparrow$}\\
    \cmidrule(lr){2-4} & Top 1 & Top 2  & Top 3 \\
    \midrule
      Real & $0.429^{\pm0.004}$ & $0.626^{\pm0.003}$ & $0.736^{\pm0.003}$ & $0.002^{\pm0.0002}$ & $3.536^{\pm0.013}$ & $9.734^{\pm0.078}$ & - \\
      \midrule
      TEMOS~\cite{petrovich22temos} & $0.092^{\pm0.003}$ & $0.171^{\pm0.003}$ & $0.238^{\pm0.002}$ & $29.258^{\pm0.0694}$ & $6.867^{\pm0.013}$ & $4.738^{\pm0.078}$ & $0.672^{\pm0.041}$ \\ 
      T2M~\cite{humanml3d} & $0.184^{\pm0.010}$ & $0.298^{\pm0.006}$ & $0.396^{\pm0.005}$ & $5.481^{\pm0.3820}$ & $9.576^{\pm 0.006}$ & $5.771^{\pm0.151}$ & $2.761^{\pm 0.042}$\\
      MDM~\cite{mdm} &  $0.203^{\pm0.009}$ & $0.329^{\pm0.007}$ & $0.426^{\pm0.005}$ & $23.701^{\pm0.0569}$ & $9.548^{\pm 0.014}$ & $5.856^{\pm0.077}$ & $3.490^{\pm0.061}$\\
      MDM(GRU)~\cite{mdm} &  $0.179^{\pm0.006}$ & $0.299^{\pm0.005}$ & $0.387^{\pm0.007}$ & $32.617^{\pm0.1221}$ & $9.557^{\pm0.019}$ & $7.003^{\pm0.134}$ & $3.430^{\pm0.035}$\\
      ComMDM~\cite{commdm} & $0.090^{\pm0.002}$ & $0.165^{\pm0.004}$ & $0.236^{\pm0.004}$ & $29.266^{\pm0.0668}$ & $\mathbf{6.870^{\pm 0.017}}$ & $4.734^{\pm0.067}$ & $0.771^{\pm 0.053}$\\
      InterGen~\cite{interhuman} & $\mathbf{0.207^{\pm0.004}}$ & $\mathbf{0.335^{\pm0.005}}$ & $\mathbf{0.429^{\pm0.005}}$ & $\mathbf{5.207^{\pm0.2160}}$ & $9.580^{\pm 0.011}$ & $\mathbf{7.788^{\pm0.208}}$ & $\mathbf{3.686^{\pm 0.052}}$\\
    \bottomrule
  \end{tabular}}
  \caption{Experimental results of text-conditioned interaction generation on the Inter-X dataset, where $\pm$ indicates 95\% confidence interval and $\rightarrow$ means the closer the better. \textbf{Bold} indicates best results.}
  \label{tab:t2m_sota}
\end{table*}

\begin{figure*}[t]
  \centering
   \includegraphics[width=1.0\linewidth]{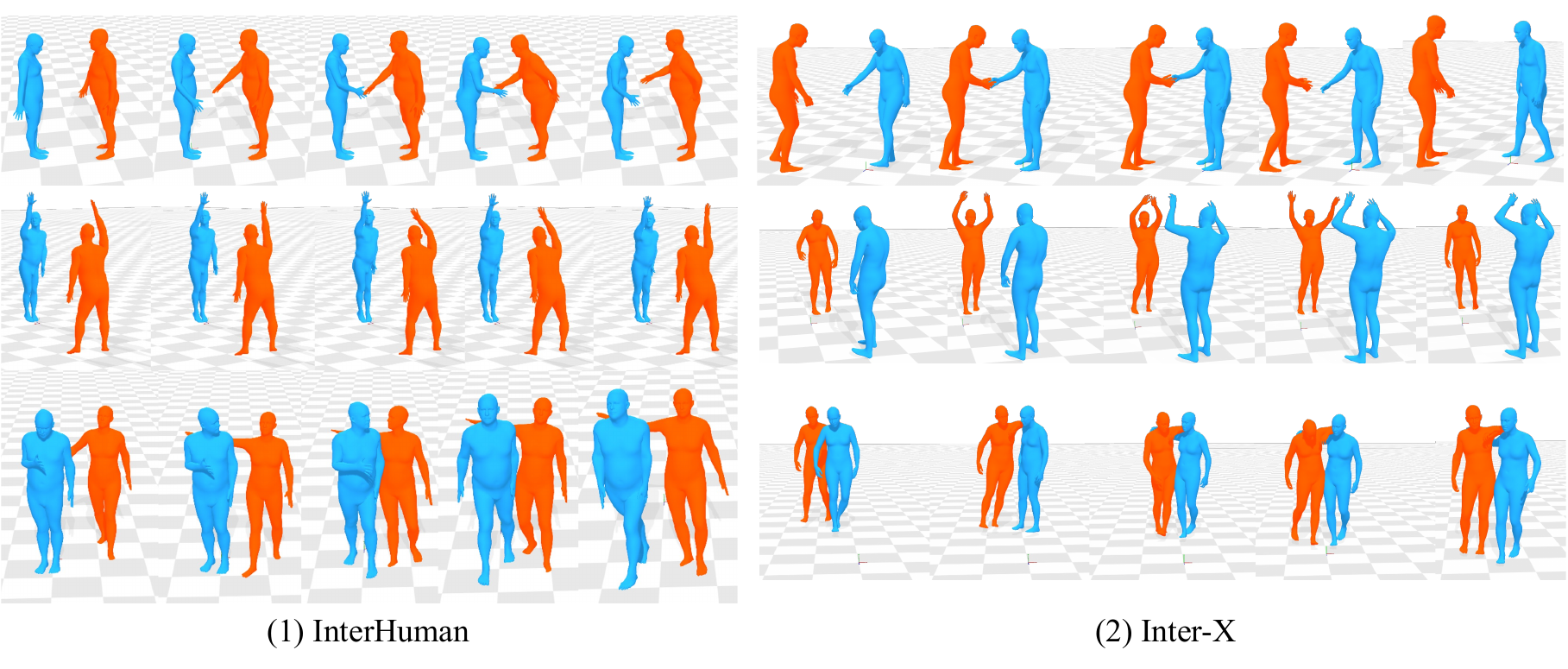}
   \caption{\textbf{Visualization results} of the generated results on the InterHuman~\cite{interhuman} and Inter-X dataset via ait-viewer~\cite{ait-viewer}. From top to bottom, the action categories are ``Handshake'', ``Wave'' and ``Shoulder to shoulder'', respectively. Please zoom in for the details.}
   \label{fig:vis}
   \vspace{-4mm}
\end{figure*}

\begin{table*}[t]
  \centering
  \begin{tabular}{@{}lcccc@{}}
    \toprule
    Method & FID$\downarrow$ & Acc.$\uparrow$ & Div.$\rightarrow$ & Multimod.$\rightarrow$\\
    \midrule
      Real & $0.281^{\pm0.002}$ & $0.990^{\pm0.0000}$ & $12.890^{\pm0.028}$ & $22.391^{\pm0.195}$ \\
      \midrule
      Action2Motion~\cite{action2motion} & $20.295^{\pm12.081}$ & $0.766^{\pm0.0003}$ & $11.581^{\pm0.024}$ & $15.345^{\pm0.245}$\\
      ACTOR~\cite{actor} & $9.392^{\pm0.816}$ & $0.855^{\pm0.0003}$ & $11.594^{\pm0.029}$ & $15.327^{\pm0.195}$ \\
      MDM~\cite{mdm} & $12.426^{\pm2.584}$ & $0.896^{\pm0.0004}$ & $13.492^{\pm0.033}$ & $\mathbf{22.042^{\pm0.153}}$\\
      MDM(GRU)~\cite{mdm} & $35.003^{\pm7.876}$ & $0.716^{\pm0.0006}$ & $\mathbf{12.579^{\pm0.038}}$ & $16.456^{\pm0.100}$\\
      Actformer~\cite{xu2023actformer} & $\mathbf{8.067^{\pm0.653}}$ & $\mathbf{0.945^{\pm0.0007}}$ & $12.512^{\pm0.05}$ & $16.187^{\pm0.189}$ \\
    \bottomrule
  \end{tabular}
  \caption{Experimental results of action-conditioned interaction generation on the Inter-X dataset. \textbf{Bold} for best results.}
  \label{tab:a2m_sota}
\end{table*}

\begin{table*}[t]
  \centering
  \begin{tabular}{@{}lcccc@{}}
    \toprule
    Method & FID$\downarrow$ & Acc.$\uparrow$ & Div.$\rightarrow$ & Multimod.$\rightarrow$\\
    \midrule
      Real & $0.260^{\pm0.0021}$ & $0.988^{\pm0.0000}$ & $12.115^{\pm0.031}$ & $21.498^{\pm0.131}$ \\
      \midrule
      MDM~\cite{mdm} & $6.747^{\pm0.3153}$ & $0.903^{\pm0.0001}$ & $12.264^{\pm0.051}$ & $19.681^{\pm0.234}$ \\
      MDM(GRU)~\cite{mdm} & $19.968^{\pm1.1700}$ & $0.752^{\pm0.0003}$ & $12.351^{\pm0.049}$ & $18.056^{\pm0.156}$\\
      RAIG~\cite{role_aware} & $6.372^{\pm0.2154}$ & $0.908^{\pm0.0001}$ & $12.330^{\pm0.060}$ & $20.071^{\pm0.299}$ \\
      AGRoL~\cite{agrol} & $\mathbf{4.386^{\pm0.2186}}$ & $\mathbf{0.925^{\pm0.0001}}$ & $\mathbf{12.204^{\pm0.042}}$ & $\mathbf{20.199^{\pm0.226}}$\\
    \bottomrule
  \end{tabular}
  \caption{Experimental results of human reaction generation based on action labels on the Inter-X dataset. \textbf{Bold} for best results.}
  \label{tab:regen_sota}
\end{table*}

\begin{table}[t]
  \centering
  \begin{tabular}{@{}lcc@{}}
    \toprule
    Method & Top-1 (\%) & Top-5 (\%) \\
    \midrule
      ST-GCN~\cite{stgcn} & 64.62 & 90.16\\
      2s-AGCN~\cite{2s-agcn} & 75.22 & 93.73 \\
      HD-GCN~\cite{hdgcn} & 77.40 & 94.73\\
      CTR-GCN~\cite{ctrgcn} & 82.19 & 96.72\\
      MS-G3D~\cite{ms-g3d} & \textbf{83.30} & \textbf{97.09}\\
    \bottomrule
  \end{tabular}
  \caption{Experimental results of skeleton-based human interaction recognition on the Inter-X dataset. \textbf{Bold} for best results.}
  \label{tab:m2a_sota}
\end{table}

\section{Experiments}

We extensively evaluate the state-of-the-art methods on the Inter-X dataset for the proposed downstream tasks with detailed discussion and analysis. In the main manuscript, we present four appealing tasks: 1) text-conditioned human interaction generation; 2) action-conditioned human interaction generation; 3) human reaction generation; and 4) human interaction recognition. The remaining experiments are presented in the supplementary materials.

\subsection{Text-conditioned Interaction Generation}

The detailed textual annotations combined with the human-human interaction sequences allow for human interaction generation. We extensively evaluate 6 state-of-the-art text to motion models, \ieno, TEMOS~\cite{petrovich22temos}, T2M~\cite{humanml3d}, MDM~\cite{mdm}, MDM-GRU~\cite{mdm,cho2014learning}, ComMDM~\cite{commdm} and InterGen~\cite{interhuman}. We modify the input and output dimensions to extend the single-person models to two-person settings and change the motion representation to SMPL-X~\cite{smplx} parameters.

\noindent\textbf{Experiment setup.}
We adopt the same protocol of~\cite{humanml3d,interhuman} to split our dataset into training, test, and validation sets with a ratio of 0.8, 0.15, and 0.05. Following~\cite{make_an_animation}, we directly borrow the SMPL-X parameters of Inter-X rather than the manually designed motion representation as in~\cite{humanml3d,interhuman}. Different from single-person motion sequences that are canonicalized to the first frame, we keep the global translation of the interacted persons so that their relative positions are reserved. For all the methods, we adopt the 6D continuous rotation representation~\cite{rot_6d} as previous works~\cite{actor,mdm,humanml3d,interhuman}. For the diffusion-based models~\cite{sohl2015deep,ddpm}, we train them with 1,000 noising timesteps and run 5 DDIM~\cite{ddim} sampling steps. Each model is trained on 4 NVIDIA A100 GPUs.

\noindent\textbf{Evaluation metrics.}
We follow~\cite{humanml3d} to adopt the Frechet Inception Distance (FID)~\cite{fid} to measure the latent distance between real and generated samples, diversity to measure latent variance, multimodality (MModality) to measure the diversity of the generated results for the same text, R Precision to measure the top-1, top-2 and top-3 accuracy of retrieving the ground-truth description from 31 randomly mismatched descriptions, and MultiModal distance (MM Dist) to calculate the latent distance between generated motions and texts. 
We train a motion feature extractor together with a text feature extractor in a contrastive manner to better align the features of texts and motions.
We run all the evaluations 20 times (except MModality for 5 times) and report the averaged results with the confidence interval at 95\%. 

\noindent\textbf{Quantitative results.}
The experimental results are depicted in~\cref{tab:t2m_sota}. We can derive that InterGen~\cite{interhuman} achieves state-of-the-art performance except for the MM Dist metric while ComMDM~\cite{commdm} achieves the worst R Precision scores. One possible explanation could be that ComMDM requires extra pre-training. From the results, we derive that our Inter-X dataset has the potential for further explorations.

\noindent\textbf{Qualitative results.}
We demonstrate the human-human interaction results generated from InterGen~\cite{interhuman} together with the generated results for the InterHuman dataset for visual comparisons in~\cref{fig:vis}. The visualization results show that with our Inter-X, the expressibility of the human-human interaction is highly enhanced with detailed hand movements. Since InterHuman does not provide dexterous hand gestures, the generated results for ``Handshake'', ``Wave'' and ``Shoulder to shoulder'' are unplausible. Besides, the synthesized results of InterHuman contain occlusions and penetrations, while ours are much more precise.

Please refer to the supplementary materials for more visual comparisons and \textbf{video} results.

\subsection{Action-conditioned Interaction Generation}

Inter-X contains 40 semantic action categories, which are currently the largest compared to other human-human interaction datasets. We conduct experiments of action-conditioned human interaction generation with the state-of-the-art methods, \ieno, Action2Motion~\cite{action2motion}, ACTOR~\cite{actor}, MDM~\cite{mdm}, MDM-GRU~\cite{mdm,cho2014learning} and Actformer~\cite{xu2023actformer}. Same as the text-conditioned methods, we re-implement these methods to adapt to our dataset format. We adopt the same dataset split protocol and pose representation as the text-conditioned methods.

\noindent\textbf{Evaluation metrics.}
Similar to the previous works~\cite{action2motion,actor,mdm} for human motion generation, we also adopt the Frechet Inception Distance (FID)~\cite{fid}, action recognition accuracy, diversity, and multi-modality for evaluation. For all these metrics, we train an action recognition model~\cite{stgcn} for feature extraction as in previous works. We generate 1,000 samples 20 times and report the average score with a confidence score of 95\%.

\noindent\textbf{Quantitative results.}
From the experimental results in~\cref{tab:a2m_sota}, Actformer~\cite{xu2023multimodal} achieves the best FID and action recognition accuracy, MDM~\cite{mdm} achieves the best Multimod. score and MDM-GRU~\cite{mdm,cho2014learning} yields the best diversity score. Although the interaction transformer is designed to model the interaction between persons, there is still substantial potential for further improvements.

\subsection{Human Reaction Generation}

We explicitly annotate the interaction order for causal human interactions, \ieno, human reaction generation. We select the MDM~\cite{mdm}, MDM-GRU~\cite{mdm,cho2014learning}, RAIG~\cite{role_aware} and AGRoL~\cite{agrol} models for evaluation. We modify the architecture of all these methods so that the motion of the actor serves as the input conditions into the model, and the output is the human reaction.

\noindent\textbf{Quantitative results.}
We demonstrate the quantitative results in~\cref{tab:regen_sota}. We observe that AGRoL~\cite{agrol} yields the best performance 
for all the evaluation metrics, while the GRU architecture achieves the worst results.

\subsection{Human Interaction Recognition}
Inter-X is built from the MoCap system with accurate 3D skeleton data. We evaluate five state-of-the-art skeleton-based action recognition models as ST-GCN~\cite{stgcn}, 2s-AGCN~\cite{2s-agcn}, HD-GCN~\cite{hdgcn}, CTR-GCN~\cite{ctrgcn} and MS-G3D~\cite{ms-g3d} and report the results of Top-1 and Top-5 recognition accuracy in~\cref{tab:m2a_sota}. Note that for simplicity, we only employed the skeleton joint stream without ensembling with bone stream and motion streams~\cite{2s-agcn,ms-g3d}.

\noindent\textbf{Quantitative results.}
From the results, we can observe that MS-G3D~\cite{ms-g3d} achieves the best Top-1 accuracy of 83.30\%, which is not satisfactory. One possible reason is that Inter-X contains dexterous hand gestures and action/reaction diversities, which would pose new challenges and opportunities for further research works.

\section{Conclusion and Limitation}

In this paper, we propose Inter-X, a large-scale human-human interaction dataset with high-precision human body movements, diverse interaction patterns, and subtle hand gestures. We also annotate Inter-X with human-part level textual descriptions from different perspectives, the semantic interaction categories, the interaction order, and the relationship and personalities of the subjects to facilitate 4 categories of downstream tasks. The qualitative and quantitative results show that Inter-X poses challenges for human-human interaction related perceptual and generative tasks.
We hope that the Inter-X dataset will promote in-depth research works on human-human interaction analysis.

\noindent{\textbf{Limitations}}.
Our work has some limitations in the following aspects:
1) \textbf{Facial expressions:} Inter-X dataset is created through an indoor MoCap venue and non-professional actors. Thus facial expressions are not involved since the correlation between expression and motion is unreliable. A possible alternative is referring to natural outdoor scenes or professional actors to explore the correlation between emotion and interactions; 
2) \textbf{Atomic interactions:} The Inter-X dataset contains 11,388 atomic human-human interaction sequences, rather than long human-human interaction sequences. We acknowledge that real-world interactions are much more complicated with longer durations and frequent transitions. However, we believe that our dataset with high precision and diversity can still serve as a cornerstone for more complicated human-human interaction analysis.

\noindent{\textbf{Boarder impacts.}} With our proposed Inter-X dataset, one can facilitate the generative models for synthesizing human-human interaction sequences given detailed textual descriptions with plenty of applications in AR/VR and gaming. For perceptual tasks of human action recognition, one can also build intelligent models for intelligent surveillance.

\noindent\textbf{Acknowledgments}: This work is supported by NSFC (62201342, 62101325), Shanghai Municipal Science and Technology Major Project (2021SHZDZX0102), NSFC under Grant 62302246 and ZJNSFC under Grant LQ23F010008.

{
    \small
    \bibliographystyle{ieeenat_fullname}
    \bibliography{main}
}

\appendix

\begingroup

\appendix
\twocolumn[
\begin{center}
\Large{\bf Inter-X: Towards Versatile Human-Human Interaction Analysis \\ **Appendix**}
\end{center}
]

\counterwithin{table}{section}
\counterwithin{figure}{section}
\setcounter{page}{1}

\section{Extra experiments}

In this section, we report the results for the remaining four settings of 1) Human interaction captioning; 2) Causal order inference; 3) Stylized human interaction generation, and 4) Personality assessment.

\subsection{Human interaction captioning}

Human interaction captioning aims to generate precise and diverse textual descriptions given the human interaction sequences. We follow~\cite{guo2022tm2t} and evaluate for motion captioning models, \ieno, RAEs~\cite{yamada2018paired}, Seq2Seq~\cite{plappert2018learning}, SeqGAN~\cite{goutsu2021linguistic} and TM2T~\cite{guo2022tm2t}. Similar to the text-conditioned interaction generation task, we simply modify the input and output dimensions to extend these models to two-person settings and also change the motion representations to SMPL-X~\cite{smplx} representations.

We follow the same protocol as text-conditioned interaction generation to split our dataset into training, testing and validation sets.
Following~\cite{guo2022tm2t}, we also adopt the R Precision and multimodal distance, together with the Bleu~\cite{bleu}, Rouge~\cite{rouge}, Cider~\cite{cider} and BertScore~\cite{bertscore} to extensively evaluate the performance of the motion captioning models.

The quantitative results are demonstrated in~\cref{tab:m2t_sota}. We can conclude that TM2T~\cite{guo2022tm2t} achieves state-of-the-art performance for all the metrics. RAEs~\cite{yamada2018paired} fails to model long-term dependencies between human-human interaction sequences and texts, thus leading to low R Precision and linguistic evaluation metrics. Seq2seq~\cite{plappert2018learning} and SeqGAN~\cite{goutsu2021linguistic} perform better than RAEs~\cite{yamada2018paired} by introducing the attention operation and the adversarial learning paradigm.

\begin{table*}[!htbp]
  \centering
  \resizebox{1\textwidth}{!}{
  \begin{tabular}{@{}lccccccccc@{}}
    \toprule
    \multirow{2}{*}{Methods}  & \multicolumn{3}{c}{R Precision$\uparrow$} & \multirow{2}{*}{MM Dist$\downarrow$}  & \multirow{2}{*}{Bleu@1$\uparrow $} & \multirow{2}{*}{Bleu@4 $\uparrow$} & \multirow{2}{*}{Rouge $\uparrow$} & \multirow{2}{*}{Cider $\uparrow$} & \multirow{2}{*}{BertScore $\uparrow$}\\
    \cmidrule(lr){2-4} & Top 1 & Top 2  & Top 3 \\
    \midrule
      \textbf{Real Desc} & 0.442 & 0.645 & 0.778 & 3.126 & - & - & - & - & - \\
      \midrule
      RAEs~\cite{yamada2018paired} & 0.094 & 0.127 & 0.245 & 7.554 & 28.6 & 9.7 & 34.1 & 25.9 & 10.2 \\
      Seq2Seq~\cite{plappert2018learning} & 0.273 & 0.436 & 0.619 & 4.285 & 53.8 & 18.5 & 45.2 & 61.9 & 27.1 \\
      SeqGAN~\cite{goutsu2021linguistic} & 0.206 & 0.398 & 0.563 & 5.447 & 45.4 & 14.1 & 36.8 & 52.3 & 21.4 \\
      TM2T~\cite{guo2022tm2t} & \textbf{0.375} & \textbf{0.583} & \textbf{0.674} & \textbf{3.493} & \textbf{56.8} & \textbf{21.6} & \textbf{48.2} & \textbf{75.5} & \textbf{32.7}\\
    \bottomrule
  \end{tabular}}
  \caption{Experimental results of human interaction captioning on the Inter-X dataset. \textbf{Bold} indicates best results.}
  \label{tab:m2t_sota}
\end{table*}

\subsection{Causal order inference}

Causal order inference aims to determine the order of the actor and the reactor in the interaction sequences. Similar to the human interaction recognition task, we adopt the models of ST-GCN~\cite{stgcn}, 2s-AGCN~\cite{2s-agcn}, HD-GCN~\cite{hdgcn}, CTR-GCN~\cite{ctrgcn} and MS-G3D~\cite{ms-g3d} as the backbone and model this problem as a binary classification task. From the quantitative results in~\cref{tab:order_ass}, we can derive that MS-G3D~\cite{ms-g3d} yields state-of-the-art performance over all the other methods. However, we found that this task is not that simple, and the performance is far from satisfactory, \ieno, only \textbf{76.8\%}.

\begin{table}[t]
  \centering
  \begin{tabular}{@{}lc@{}}
    \toprule
    Method & Accuracy (\%)\\
    \midrule
      ST-GCN~\cite{stgcn} & 62.3\\
      2s-AGCN~\cite{2s-agcn} & 68.2 \\
      HD-GCN~\cite{hdgcn} & 70.6\\
      CTR-GCN~\cite{ctrgcn} & 74.5\\
      MS-G3D~\cite{ms-g3d} & \textbf{76.8}\\
    \bottomrule
  \end{tabular}
  \caption{Experimental results of causal order inference on the Inter-X dataset. \textbf{Bold} for best results.}
  \label{tab:order_ass}
\end{table}

\subsection{Stylized human interaction generation}
We implement the stylized human interaction generation based on the vanilla human interaction generations models, \ieno, Action2Motion~\cite{action2motion}, ACTOR~\cite{actor}, MDM~\cite{mdm}, MDM-GRU~\cite{mdm,cho2014learning} and Actformer~\cite{xu2023actformer}. We add the familiarity level as a style code injected into the model as in~\cite{aberman2020unpaired}. We also report the Frechet Inception Distance (FID)~\cite{fid}, action recognition accuracy, diversity, and multi-modality in~\cref{tab:style_gen}. From~\cref{tab:style_gen}, we can derive that Actformer~\cite{xu2023actformer} achieves the best FID score and Accuracy, and MDM~\cite{mdm} achieves the best Diversity and Multimodality score.

\begin{table*}[t]
  \centering
  \begin{tabular}{@{}lcccc@{}}
    \toprule
    Method & FID$\downarrow$ & Acc.$\uparrow$ & Div.$\rightarrow$ & Multimod.$\rightarrow$\\
    \midrule
      Real & $0.281^{\pm0.002}$ & $0.990^{\pm0.0000}$ & $12.890^{\pm0.028}$ & $22.391^{\pm0.195}$ \\
      \midrule
      Action2Motion~\cite{action2motion} & $21.182^{\pm13.319}$ & $0.737^{\pm0.0005}$ & $11.492^{\pm0.032}$ & $14.934^{\pm0.258}$ \\
      ACTOR~\cite{actor} & $9.796^{\pm0.862}$ & $0.867^{\pm0.0003}$ & $11.862^{\pm0.039}$ & $15.174^{\pm0.245}$\\
      MDM~\cite{mdm} & $11.762^{\pm1.854}$ & $0.912^{\pm0.0002}$ & $\mathbf{13.025^{\pm0.028}}$ & $\mathbf{21.742^{\pm0.106}}$\\
      MDM(GRU)~\cite{mdm} & $31.688^{\pm4.492}$ & $0.753^{\pm0.0006}$ & $12.259^{\pm0.039}$ & $16.271^{\pm0.206}$\\
      Actformer~\cite{xu2023actformer} & $\mathbf{8.544^{\pm0.684}}$ & $\mathbf{0.932^{\pm0.0006}}$ & $12.116^{\pm0.062}$ & $16.122^{\pm0.183}$\\
    \bottomrule
  \end{tabular}
  \caption{Experimental results of action-conditioned stylized human interaction generation on the Inter-X dataset. \textbf{Bold} for best results.}
  \label{tab:style_gen}
\end{table*}

\subsection{Personality assessment}
Personality assessment is to automatically obtain personalities through human interactions. Different from the previous dataset splitting methods, we split the train/test/val sets by person IDs with the ratio of 0.8, 0.15 and 0.05. We also adopt the models of ST-GCN~\cite{stgcn}, 2s-AGCN~\cite{2s-agcn}, HD-GCN~\cite{hdgcn}, CTR-GCN~\cite{ctrgcn} and MS-G3D~\cite{ms-g3d} as the backbone and model this problem as a regression task. We report the $R^2$ values for each personality element. From the quantitative results in~\cref{tab:per_ass}, we can derive that MS-G3D~\cite{ms-g3d} achieves the best performance over all the other methods, except for the element of ``Agreeableness'', and CTR-GCN~\cite{ctrgcn} achieves the best $R^2$ score for the ``Agreeableness''.

\begin{table*}[t]
  \centering
  \resizebox{0.8\textwidth}{!}{
  \begin{tabular}{@{}lccccc@{}}
    \toprule
    Method & Openness & Conscientiousness & Extraversion & Agreeableness & Neuroticism \\
    \midrule
      ST-GCN~\cite{stgcn} & 21.16 & 25.38 & 34.91 & 23.67 & 13.02 \\
      2s-AGCN~\cite{2s-agcn} & 23.46 & 31.27 & 38.72 & 24.88 & 13.57 \\
      HD-GCN~\cite{hdgcn} & 25.92 & 33.19 & 41.33 & 26.83 & 14.29\\
      CTR-GCN~\cite{ctrgcn} & 27.78 & 35.41 & 43.52 & \textbf{29.43} & 15.63\\
      MS-G3D~\cite{ms-g3d} & \textbf{28.36} & \textbf{37.88} & \textbf{46.23} & 29.07 & \textbf{16.35}\\
    \bottomrule
  \end{tabular}}
  \caption{The $R^2$ values results (\%) of the personality assessment on the Inter-X dataset. \textbf{Bold} for best results.}
  \label{tab:per_ass}
\end{table*}

\begin{table*}[t]
  \centering
  \begin{tabular}{@{}|l|l|l|l|@{}}
    \toprule
    A01: Hug & A02: Handshake & A03: Wave & A04: Grab \\
    \midrule
    A05: Hit & A06: Kick & A07: Posing & A08: Push \\
    \midrule
    A09: Pull & A10: Sit on leg & A11: Slap & A12: Pat on back \\
    \midrule
    A13: Point finger at & A14: Walk towards & A15: Knock over & A16: Step on foot \\
    \midrule
    A17: High-five & A18: Chase & A19: Whisper in ear & A20: Support with hand \\
    \midrule
    A21: Finger-guessing & A22: Dance & A23: Link arms & A24: Shoulder to shoulder \\
    \midrule
    A25: Bend & A26: Carry on back & A27: Massage shoulder & A28: Massage leg \\
    \midrule
    A29: Hand wrestling & A30: Chat & A31: Pat on cheek & A32: Thumb up \\
    \midrule
    A33: Touch head & A34: Imitate & A35: Kiss on cheek & A36: Help up \\
    \midrule
    A37: Cover mouth & A38: Look back & A39: Block & A40: Fly kiss \\
    \bottomrule
  \end{tabular}
  \caption{The action categories of Inter-X.}
  \label{tab:actions}
\end{table*}

\section{SMPL-X optimization details}

Formally, our SMPL-X parameters consist of the body pose parameters $\theta \in \mathbb{R}^{N\times55\times 3}$, translation $t\in \mathbb{R}^{N\times3}$ and the shape parameters $\beta\in \mathbb{R}^{N\times10}$, where $N$ is the number of frames. We initialize the subjects' shape $\beta$ based on their height and weight as~\cite{virtual_caliper}. Then a two-stage SMPL-X optimization algorithm is adopted to our Mocap data to obtain the SMPL-X parameters.

In the first stage, we only optimize the pose parameters except that of fingers. The joint energy term 
\begin{equation}
    \mathbb{E}_j=\frac{1}{N}\sum\limits_{i=0}\limits^{N}\sum\limits_{j\in\mathcal{J}}\|\bm{J}_j^i(\mathbb{M}(\theta_b,t)-\bm{g}_j^i\|_2^2
\end{equation}
aims to fit the SMPL-X joints to our captured skeleton data, where $\mathcal{J}$ denotes the joint set, $\mathbb{M}$ is the SMPL-X parametric model, $\bm{J}_j^i$ is the joint regressor function for joint $j$ at $i$-th frame, $\theta_b$ is the pose parameters excluding fingers, $\bm{g}$ is the Mocap skeleton data. A smoothing term 
\begin{equation}
    \mathbb{E}_{smooth}=\frac{1}{N-1}\sum\limits_{i=0}\limits^{N-1}\sum\limits_{j\in\mathcal{J}}\|\bm{J}_j^{i+1}-\bm{J}_j^{i}\|_2^2
\end{equation}
alleviates the pose jittering between frames. A regularization term 
\begin{equation}
    \mathbb{E}_{r}=\|\theta_b\|_2^2
\end{equation}
constrains the pose parameters from deviating too much. In total, our optimization objective at the first stage is:
\begin{equation}
    \mathbb{E}_1=\lambda_j\mathbb{E}_j+\lambda_{smooth}\mathbb{E}_{smooth}+\lambda_r\mathbb{E}_r,
\end{equation}
and we set $\lambda_j,\lambda_{smooth},\lambda_r=1,0.1,0.01$.

For the second stage, we append the finger pose parameters and jointly optimize the whole-body pose parameters. We especially emphasize fingers' optimization, thus we separate fingers' pose parameters from the body part. Our optimization objective in the second stage is summarized as:
\begin{gather}
    \mathbb{E}_b=
    \lambda_j\mathbb{E}_j+\lambda_{smooth}\mathbb{E}_{smooth}+\lambda_r\mathbb{E}_r, \\
    \mathbb{E}_{h}=\lambda_{j_{h}}\mathbb{E}_{j_{h}}+\lambda_{{smooth}_h}\mathbb{E}_{{smooth}_h}+\lambda_{r_h}\mathbb{E}_{r_h}, \\
    \mathbb{E}_2=\mathbb{E}_b+\mathbb{E}_h,
\end{gather}
we set $\lambda_j,\lambda_{smooth},\lambda_r=1,0.1,0.01$ for the body part and $\lambda_{j_h},\lambda_{{smooth}_h},\lambda{r_h}=10,0.01,0.001$ for fingers.

\section{The action categories}

We provide the names of the 40 human-human interaction categories in~\cref{tab:actions}.

\section{Samples of textual annotations}
We provide some samples of the textual annotations of our Inter-X dataset in~\cref{fig:text_annot}.

\section{More visualization results}
We provide the rendered RGB frames based on the Unreal Engine in~\cref{fig:ue}. We also provide more visualization samples of Inter-X in the supplementary video.

\begin{figure*}[t]
  \centering
   \includegraphics[width=1.0\linewidth]{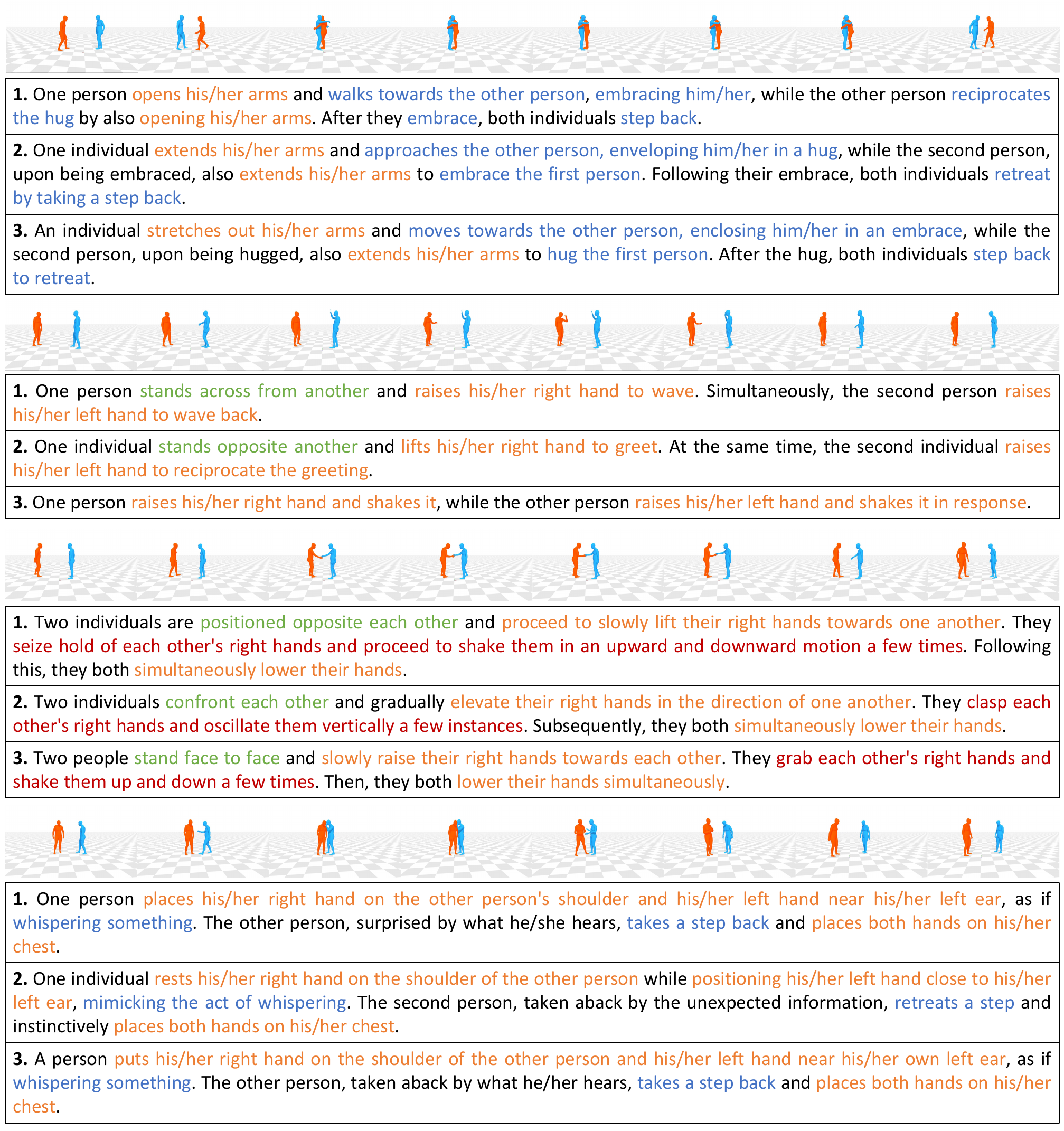}
   \caption{Some samples of the textual annotations of the Inter-X dataset.}
   \label{fig:text_annot}
   \vspace{-4mm}
\end{figure*}

\begin{figure*}[t]
  \centering
   \includegraphics[width=1.0\linewidth]{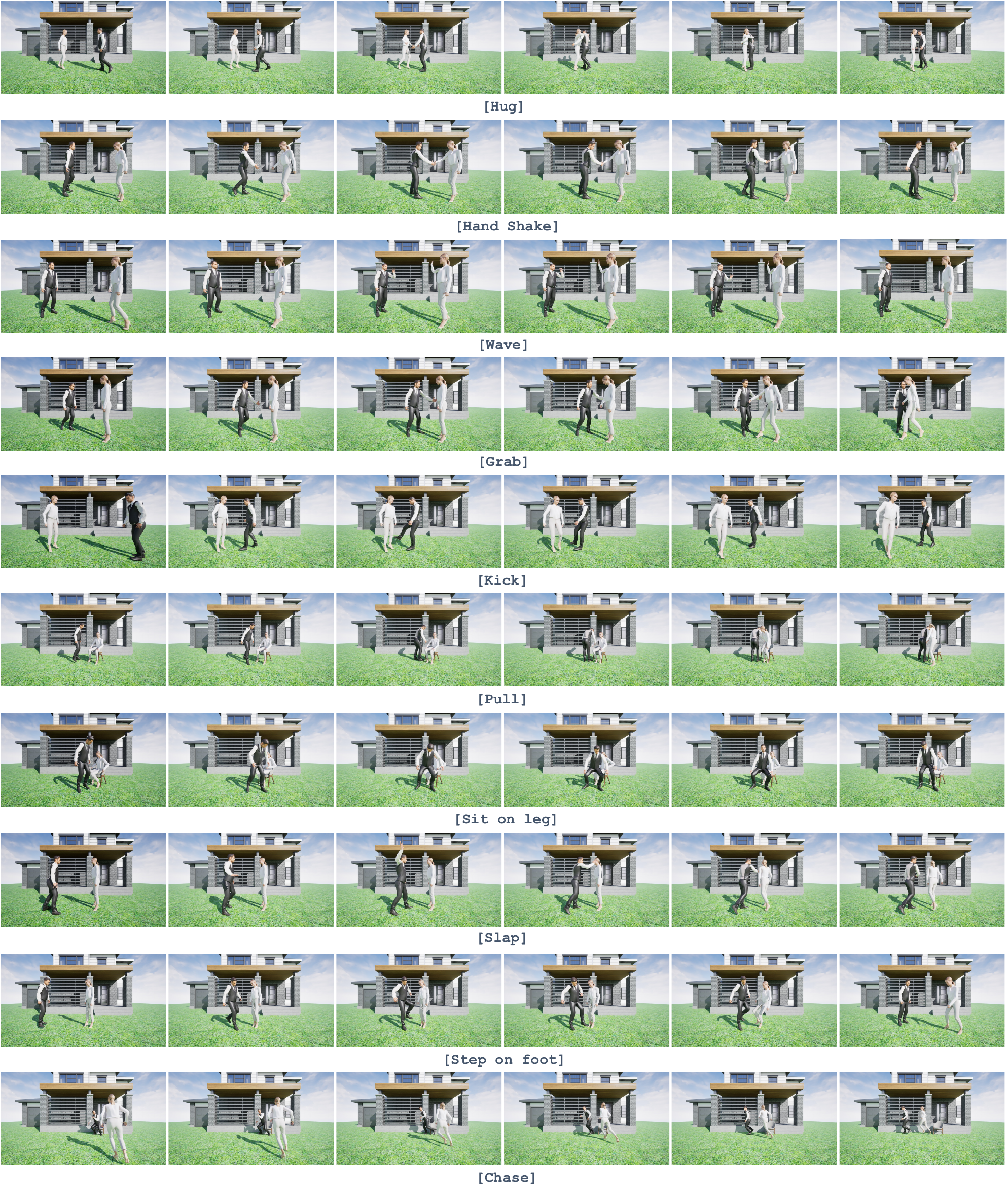}
   \caption{The visualization results of the rendered RGB frames based on the Unreal Engine.}
   \label{fig:ue}
   \vspace{-4mm}
\end{figure*}

\end{document}